\documentclass{article}
\usepackage{spconf,amsmath,graphicx}
\usepackage{multirow}
\usepackage{array}
\usepackage{xcolor}

\usepackage[utf8]{inputenc}
\usepackage{enumitem}
\usepackage{microtype}      
\usepackage{booktabs}
\usepackage{times}
\usepackage{epsfig}
\usepackage{amssymb}
\usepackage{bm}
\usepackage[bottom]{footmisc}
\usepackage{adjustbox}
\usepackage[normalem]{ulem}

\title{PromptMAD: Cross-Modal Prompting for Multi-Class Visual Anomaly Localization}

\name{Duncan McCain, Hossein Kashiani, Fatemeh Afghah\thanks{This material is based upon work supported by the National Science Foundation under Grant Numbers CNS-2232048, and CNS-2204445.\\© 2026 IEEE. Personal use of this material is permitted. Permission from IEEE must be obtained for all other uses, in any current or future media, including reprinting/republishing this material for advertising or promotional purposes, creating new collective works, for resale or redistribution to servers or lists, or reuse of any copyrighted component of this work in other works. }}

\address{
Holcombe Department of Electrical and Computer Engineering\\  Clemson University
}

\begin{document}
\topmargin=0mm
\maketitle

\begin{abstract}

Visual anomaly detection in multi-class settings poses significant challenges due to the diversity of object categories, the scarcity of anomalous examples, and the presence of camouflaged defects. In this paper, we propose \textit{PromptMAD}, a cross-modal prompting framework for unsupervised visual anomaly detection and localization that integrates semantic guidance through vision-language alignment. By leveraging CLIP-encoded text prompts describing both normal and anomalous class-specific characteristics, our method enriches visual reconstruction with semantic context, improving the detection of subtle and textural anomalies. To further address the challenge of class imbalance at the pixel level, we incorporate Focal loss function, which emphasizes hard-to-detect anomalous regions during training. Our architecture also includes a supervised segmentor that fuses multi-scale convolutional features with Transformer-based spatial attention and diffusion iterative refinement, yielding precise and high-resolution anomaly maps. Extensive experiments on the MVTec-AD dataset demonstrate that our method achieves state-of-the-art pixel-level performance, improving mean AUC to 98.35\% and AP to 66.54\%, while maintaining efficiency across diverse categories.

\end{abstract}


\section{Introduction}
\label{sec:intro}

Unsupervised visual anomaly detection (AD) is crucial in applications such as industrial defect inspection, enabling models to identify deviations from normality without access to labeled anomalous samples during training. The MVTec-AD dataset \cite{mvtec}, which spans a variety of objects and textures, serves as a key benchmark for unified AD methods that process multiple classes using a single model. Recent progress in unified AD, exemplified by approaches leveraging transformer-based architectures \cite{you2022uniad, onenip}, vision-language models \cite{jeong2023winclip}, and prompt-guided reconstruction \cite{onenip, you2022uniad,ROADS,li2024promptad}, has advanced multi-class handling. However, these methods often struggle with accurate pixel-level localization, particularly for camouflaged anomalies that blend seamlessly with their surroundings.

These limitations stem from reliance on unimodal visual cues that neglect semantic domain knowledge, challenges particularly pronounced in industrial quality control, where precise defect localization is essential. To address them, we propose PromptMAD, a text-guided diffusion model for unified anomaly detection that leverages a single normal image prompt to guide bidirectional feature reconstruction and restoration, incorporating: (1) a supervised segmentor with diffusion processes for iterative anomaly map refinement, (2) cross-modal prompting via CLIP \cite{clip} that enriches visual inputs with textual descriptions of class-specific normality and anomaly types, functioning akin to retrieval-augmented generation by providing domain-specific semantic guidance (e.g., in the transistor category: descriptions of defect types such as bent lead, cut lead, damaged case, misplaced, alongside a normal case; for the pill category: color anomalies, combined defects, contamination, crack, scratch, faulty imprint, alongside defect-free pill), and (3) Focal loss function to emphasize hard-to-detect anomalous pixels amid class imbalance. These innovations enable accurate reconstruction of normal features while highlighting anomalies, promoting object-agnostic generalization across classes without compromising efficiency in resource-constrained settings. Comprehensive experiments on MVTec-AD \cite{mvtec} demonstrate substantial improvements in pixel-level metrics.\vspace{-1mm}

\begin{figure*}[!t]
    \centering
    \includegraphics[width=\linewidth]{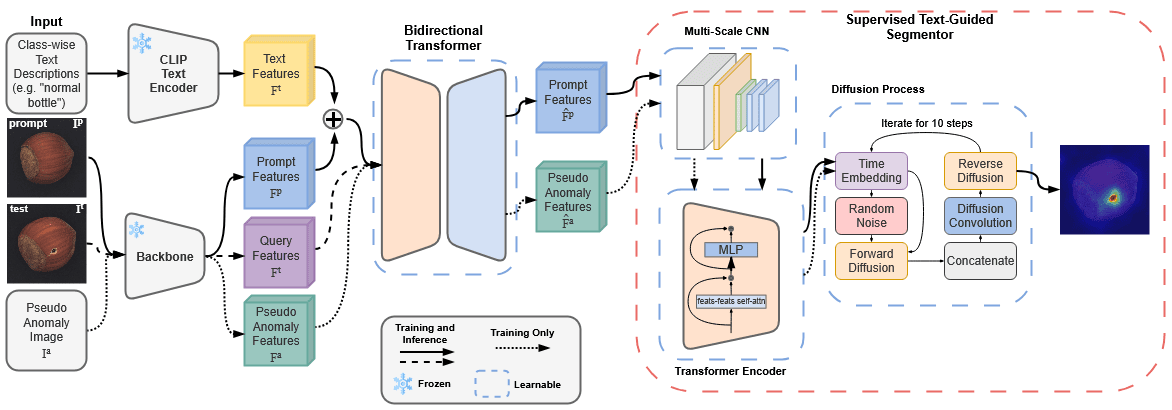}
    \caption{Overview of PromptMAD architecture for multi-class visual anomaly localization. A pre-trained backbone extracts multi-scale features from target and normal prompt images. Cross-modal prompts fuse visual tokens with CLIP-encoded textual descriptions of normal and anomalous class traits (e.g., "uniform threads, no deformations" vs. "bent leads, cracks") for semantic guidance. The bidirectional Transformer decoder reconstructs normal features while amplifying anomalies. Reconstruction errors feed into an enhanced supervised refiner with deeper CNNs for multi-scale extraction, Transformer attention for spatial focus, and diffusion-based iterative refinement for high-resolution anomaly maps.}
    \label{fig:architecture}\vspace{-2mm}
\end{figure*}

\section{RELATED WORKS}
\label{sec:relatedworks}

\noindent{\textbf{Unsupervised Anomaly Detection.}} Embedding-based methods \cite{defard2021padim,rippel2021MDND} leverage pre-trained features for anomaly scoring via distances or memory banks. Discriminator-based approaches \cite{draem, zhang2023prn} generate pseudo-anomalies (e.g., via CutPaste \cite{li2021cutpaste}) to enable supervised-like training. Reconstruction-based techniques \cite{mou2023rgi, ZAVRTANIK2021riad} exploit poor anomaly reconstruction from normal-trained autoencoders or GANs. Recent prompt-based methods \cite{onenip, you2022uniad,ROADS,li2024promptad} enhance reconstruction with contextual guidance, effectively handling both overt and camouflaged anomalies.\vspace{1mm}

\noindent{\textbf{Prompt-Based and Multimodal Anomaly Detection.}} 
Vision-language models (VLMs) such as CLIP \cite{clip} enable zero- and few-shot anomaly detection via prompt engineering and prompt learning \cite{talemi2025style,alipour2025disa}. Early works use hand-crafted textual prompts \cite{jeong2023winclip}, while later methods learn prompts during training \cite{li2024promptad, zhou2024anomalyclip}. However, existing approaches often underexploit multimodal synergy. We address this by fusing visual prompts with semantic text descriptions, leveraging CLIP alignment for improved generalization beyond single-modality constraints. \vspace{1mm}

\noindent{\textbf{{Loss Functions and Refiners in Anomaly Detection.}}
Imbalanced anomaly segmentation benefits from losses like Dice \cite{Sudre_2017dice} or Focal loss function \cite{lin2017focal}, which emphasize challenging samples. Refiners \cite{draem} upscale error maps, but shallow designs limit precision. We extend these with Transformer-inspired \cite{attention} attention and diffusion models \cite{ho2020ddpm} for iterative enhancement, hypothesizing that such depth captures long-range dependencies for superior localization.\vspace{-1mm}

\section{METHOD}
\label{sec:method}
Our proposed text-guided diffusion model for unified anomaly detection builds upon prompt-based reconstruction techniques to address key limitations in precise pixel-level localization: class imbalance in segmentation and reliance on unimodal visual guidance. We adopt OneNIP \cite{onenip} as the baseline framework owing to its state-of-the-art efficacy in multi-class anomaly detection and its modular design, which enables targeted enhancements while preserving computational efficiency.

\subsection{Baseline Overview}
\label{ssec:onenip_overview}

The baseline framework \cite{onenip} employs a single normal image prompt to guide bidirectional feature reconstruction and restoration in a prompt-based anomaly detection paradigm. A pre-trained backbone, EfficientNet-B4 \cite{tan2019efficientnet}, extracts multi-level features from the input target image and the prompt, which are subsequently tokenized and fed into a transformer-based bidirectional decoder. We bilinearly resize backbone stages to $14\times 14$ and concatenate them into a 272-channel feature map before tokenization. The decoder utilizes self-attention to capture contextual dependencies and cross-attention to incorporate prompt-driven guidance, yielding reconstruction error maps that delineate deviations from normality. These error maps are upscaled to high-resolution anomaly masks using a supervised segmentor trained on synthetic pseudo-anomalies generated via techniques like CutPaste. The overall training objective integrates mean squared error (MSE) loss to enforce reconstruction fidelity and Dice loss to optimize segmentation accuracy, enabling effective pixel-level anomaly detection during inference. Although this baseline excels in unified multi-class settings, it is limited by uniform pixel error handling and exclusive reliance on visual prompts, which our text-guided enhancements address without increasing inference overhead. PromptMAD extends this foundation with semantic cross-modal prompts, and diffusion-guided refinement for improved anomaly localization.

\subsection{Class Imbalance Mitigation}
\label{ssec:focallossintegration}
Anomaly segmentation suffers from pronounced class imbalance, with normal pixels overwhelmingly dominating sparse anomalous regions. Unlike prior reconstruction-based methods \cite{draem, you2022uniad} that rely on MSE and Dice losses, which treat all pixels uniformly and thus underemphasize rare anomalies, we incorporate the Focal loss function \cite{lin2017focal} into the training objective as \(\mathcal{L}_{\text{focal}} = \alpha (1 - p_t)^\gamma \mathcal{L}_{\text{BCE}}(p, y)\), where \(p_t\) is the predicted probability for the ground-truth class, \(\alpha\) balances class weights, \(\gamma\) modulates focus on difficult examples, and \(\mathcal{L}_{\text{BCE}}\) denotes binary cross-entropy. This modulating factor \((1 - p_t)^\gamma\) downweights well-classified (easy) pixels while amplifying losses for challenging, low-confidence predictions, thereby prioritizing anomalous regions in the text-guided diffusion process.

\subsection{Text-Guided Segmentor}

To achieve finer anomaly localization, we introduce a text-guided supervised segmentor that integrates a deeper convolutional model, transformer-based spatial attention, and diffusion-based iterative refinement for multi-scale error map enhancement. The convolutional component adopts a multi-scale residual architecture \cite{he2016resnet}, comprising a stem for initial downsampling, stacked layers with batch normalization and ReLU, and skip connections for fusing hierarchical features. This enables robust extraction of anomaly patterns at varied resolutions from high-dimensional inputs. The transformer encoder \cite{dosovitskiy2021vit} models long-range dependencies by reshaping the input features into patch sequences of shape $[B, H \times W, C]$, where $B$ denotes the batch size, $H \times W$ is the total number of patches, and $C$ is the number of channels. The encoder then applies multi-head self-attention followed by layer normalization. This bolsters segmentation of subtle anomalies by incorporating global context. Finally, the diffusion module employs a denoising process \cite{ho2020ddpm, zhang2023diffboost} to iteratively refine the initial anomaly map, computed as the $L_1$ difference between reconstructed and query features, over 10 steps with a linear \texttt{beta} schedule (from \(10^{-4}\) to 0.02). Time embeddings condition convolutional denoising, complemented by U-Net-like upsampling \cite{ronneberger2015unet} for high-resolution outputs. The diffusion refiner is a lightweight 2D U-Net denoiser with sinusoidal timestep embeddings (2-layer MLP) injected into residual blocks and fused with the CLIP text embedding. Conditioned on CLIP-enriched text prompts describing class-specific defects, this text-guided diffusion elevates boundary precision and average precision in segmentation.

\subsection{Cross-Modal Prompt Enrichment with CLIP}
\label{ssec:prompt_enhancement}

The bidirectional decoder processes tokenized features from the target image alongside visual prompts derived from a single normal reference. To infuse semantic guidance, we augment these with class-specific textual descriptions via frozen CLIP \cite{clip}: texts detailing normality and anomaly types (e.g., for transistor: normal case alongside defects like bent lead, cut lead, damaged case, and misplaced) are encoded using CLIP's text encoder, projected to 256 dimensions, and fused element-wise with visual prompt embeddings before decoder input. This cross-modal integration exploits CLIP's pre-trained vision-language alignment to enhance contextual reconstruction, fostering improved generalization to camouflaged and textured anomalies without fine-tuning the visual backbone or increasing latency.

\section{EXPERIMENTS}
\label{sec:experiments}

\subsection{Dataset and Setup}
\label{ssec:dataset_setup}

We evaluate PromptMAD on the MVTec Anomaly Detection (MVTec-AD) dataset \cite{mvtec}, which includes 15 categories comprising 10 object classes and 5 texture classes. The training set consists of 3,629 normal images, while the test set contains 1,725 images with both normal and anomalous samples. All images are resized to $224 \times 224$ pixels. Models are trained for 1,500 epochs using the AdamW optimizer with a learning rate of $1\text{e}^{-4}$, weight decay $1\text{e}^{-5}$, a StepLR learning rate scheduler, and Focal loss function configured with $\alpha = 0.75$ and $\gamma = 2$. Performance is evaluated using both image-level and pixel-level metrics. For both levels, we report the Area Under the ROC Curve (AUC) and Average Precision (AP), following the evaluation protocol from \cite{onenip}. We compare our results against the OneNIP baseline. All experiments are implemented in PyTorch 2.4.1 with CUDA 12.4 and run on 4 NVIDIA A100 GPUs.

\begin{figure}[!t]
    \centering
    \includegraphics[width=\linewidth]{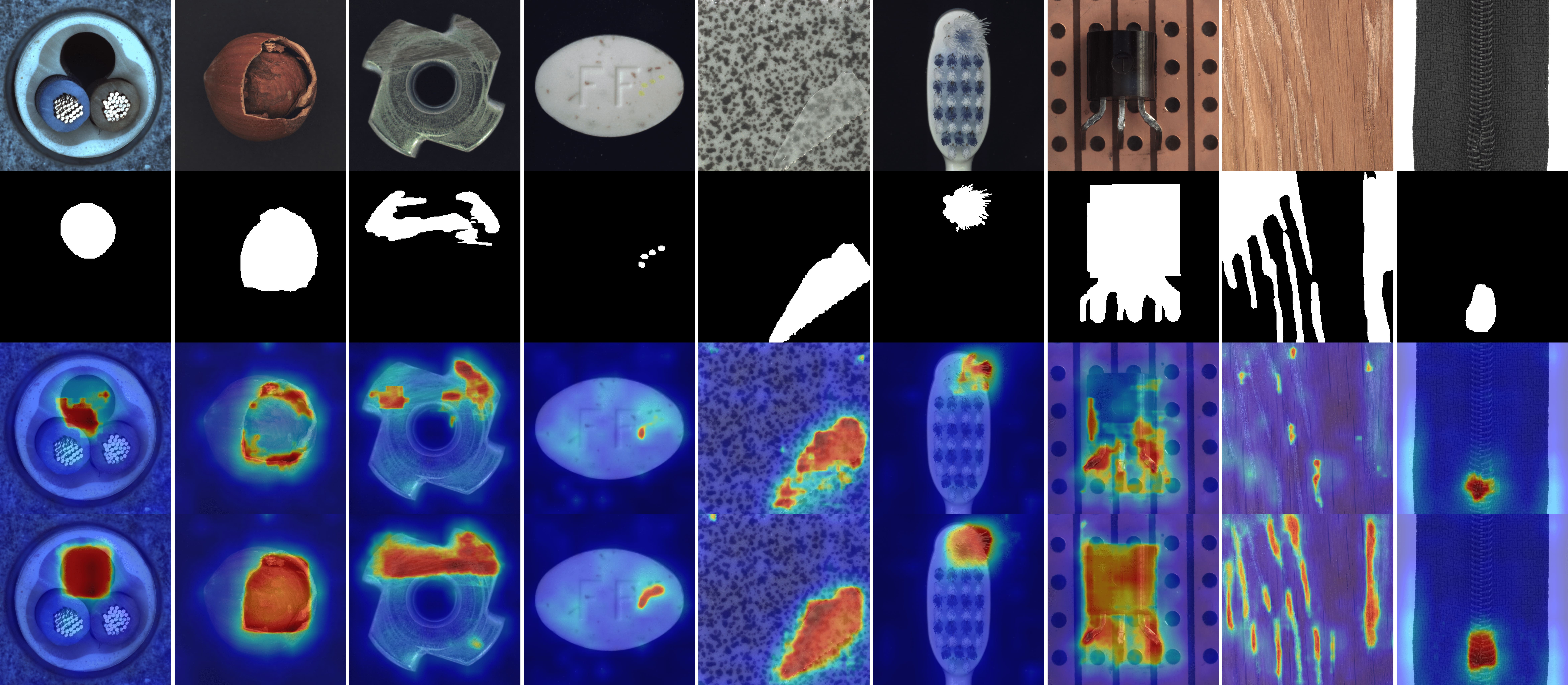}
    \caption{Qualitative comparison of anomaly localization on selected MVTec-AD test examples (cable, hazelnut, metal nut, pill, tile, toothbrush, transistor, wood, zipper). From top to bottom: input anomalous images, ground truth masks, OneNIP \cite{onenip} anomaly maps, our PromptMAD anomaly maps.}
    \label{fig:results}
\end{figure}

\subsection{Main Results}
\label{ssec:main_results}

\begin{table}[]
    \centering
    \caption{Image-level anomaly classification and pixel-level anomaly segmentation comparisons with AUC/AP metrics on MVTec-AD dataset \cite{mvtec}. All methods are evaluated under the unified setting. The best results are highlighted in bold.}
    \vspace{1mm} 
    \label{tab:comparison}
    \resizebox{\linewidth}{!}{
        \begin{tabular}{|c|c|c|c|c|c|}
        \hline
        Metric   & DRAEM \cite{draem} & UniAD \cite{you2022uniad} & OneNIP \cite{onenip} & PromptMAD \\
        \hline
        I-AUC/AP & 91.4 / 95.3 & 96.5 / 98.9  & 97.2 / 99.1 & \textbf{97.6} / \textbf{99.2} \\
        P-AUC/AP & 85.2 / 49.6 & 96.8 / 44.7  & 97.8 / 63.5 & \textbf{98.3} / \textbf{66.5} \\
        \hline
        \end{tabular}
    }
\end{table}

Table \ref{tab:onenip_results} compares our PromptMAD against the baseline across all 15 MVTec-AD classes. Our method achieves significant improvements in pixel-level metrics: mean pixel AUC increases from 97.81\% to 98.35\% (+0.54\%) and pixel AP from 63.52\% to 66.54\% (+3.02\%). Image-level performance shows modest gains with max AUC improving from 97.20\% to 97.62\% and max AP from 99.07\% to 99.23\%. Per-class analysis highlights significant gains in challenging categories, particularly texture-rich classes like pill, metal\_nut, hazelnut, and tile, which benefit from cross-modal prompting that leverages CLIP-encoded textual descriptions to distinguish defects from normal variations. Some classes exhibit trade-offs: capsule shows improved AUC but minimal pixel AP change, while grid and screw experience slight pixel AP reductions due to Focal loss function prioritizing harder examples.

The enhanced segmentor, incorporating multi-scale convolutional and transformer-based attention alongside diffusion-based refinement, ensures consistent pixel AUC improvements across all classes. Focal loss function further enhances detection of sparse anomalies, notably in bottle (+2.5\% pixel AP) and wood (+5.4\% pixel AP). These results underscore the efficacy of our cross-modal prompting and text-guided diffusion approach in achieving precise localization of camouflaged defects in industrial settings. To ensure our enhancements preserve efficiency, we measured inference runtimes on 4 NVIDIA A100 GPUs for MVTec-AD test samples. OneNIP achieves 219 FPS (4.5 ms per sample), while PromptMAD, despite added components, records 193 FPS (5.2 ms per sample), maintaining real-time suitability for industrial applications with improved accuracy.

\subsection{Qualitative Analysis}
\label{ssec:qualitativeanalysis}

Figure \ref{fig:results} illustrates anomaly localization comparisons between the baseline and our text-guided method across ten MVTec-AD classes. Our approach yields sharper anomaly boundaries and fewer false positives, especially in textured categories like tile and wood, where the baseline misidentifies normal variations as defects. The enhanced segmentor, integrating deeper convolutional layers and transformer attention, effectively captures multi-scale anomalies, as demonstrated by superior localization of small defects (e.g., in pill and zipper) and extended regions (e.g., in cable, hazelnut, and toothbrush). Cross-modal prompting further aids in differentiating genuine defects from texture artifacts, evident in examples such as metal nut, tile, and wood. Moreover, Focal loss function integration bolsters detection of camouflaged anomalies (e.g., in pill and transistor) by prioritizing challenging samples during training. These qualitative enhancements corroborate the observed quantitative improvements in pixel-level metrics.

\subsection{Ablation Study}
To assess the impact of our components, we conduct ablations on the MVTec-AD dataset \cite{mvtec} in Table \ref{tab:ablation}, systematically removing elements while maintaining consistent training (1,500 epochs, AdamW optimizer with learning rate $10^{-4}$). Comparisons employ image-level (maximum AUC and AP) and pixel-level (pixel AUC and AP) metrics. The baseline achieves a mean pixel AUC of 97.81\% and pixel AP of 63.52\%. Employing only the proposed segmentor results in a pixel AUC of 97.53\% and pixel AP of 62.14\%, yielding gains in bottle (84.22\% vs. 82.62\%) and toothbrush (59.76\% vs. 52.94\%) through refined boundaries, though declines in screw (26.97\% vs. 37.79\%) underscore imbalance challenges.

Cross-modal prompting alone produces a pixel AUC of 97.70\% and pixel AP of 62.49\%, favoring textured classes such as carpet (68.58\% vs. 68.39\%) and zipper (59.36\% vs. 59.04\%) via semantic enrichment, yet reducing performance in object classes like metal nut (73.24\% vs. 77.66\%). Focal loss function in isolation yields a pixel AUC of 97.53\% and pixel AP of 62.19\%, boosting sparse anomalies in bottle (75.53\% vs. 72.92\%) and wood (71.23\% vs. 67.71\%), with slight image-level reductions (maximum AUC: 96.02\% vs. 97.20\%).

\begin{table}[t]
    \centering
    \caption{Comparison of our proposed method with baseline OneNIP on MVTec-AD dataset \cite{mvtec}. Metrics include image-level (Max AUC, Max AP) and pixel-level (Pixel AUC (P-AUC), Pixel AP (P-AP)) performance. Our method improves mean pixel-level metrics, with notable gains in classes like hazelnut and tile.}
    \label{tab:onenip_results}
    \vspace{1mm} 
    \resizebox{\linewidth}{!}{
        \begin{tabular}{|l|c|c|c|c||c|c|c|c|}
        \hline
        \multirow{2}{*}{Class} & \multicolumn{4}{c||}{OneNIP \cite{onenip}} & \multicolumn{4}{c|}{PromptMAD} \\
        \cline{2-9}
        & AUC & AP & P-AUC & P-AP & AUC & AP & P-AUC & P-AP \\
        \hline
        Bottle     & 99.84 & 99.95 & 98.61 & 82.62 & 100.00 & 100.00 & 99.01 & 85.13 \\
        Cable      & 97.53 & 98.54 & 97.94 & 63.48 & 97.94 & 98.66 & 98.02 & 65.01 \\
        Capsule    & 85.08 & 96.31 & 98.48 & 50.06 & 88.59 & 97.13 & 98.50 & 48.88 \\
        Carpet     & 99.68 & 99.90 & 98.94 & 68.39 & 99.20 & 99.77 & 98.88 & 70.31 \\
        Grid       & 98.83 & 99.65 & 98.12 & 45.43 & 99.16 & 99.74 & 98.34 & 42.31 \\
        Hazelnut   & 100.00 & 100.00 & 98.80 & 72.92 & 100.00 & 100.00 & 99.46 & 82.45 \\
        Leather    & 100.00 & 100.00 & 99.59 & 71.07 & 100.00 & 100.00 & 99.67 & 71.62 \\
        Metal Nut  & 99.51 & 99.89 & 97.11 & 77.66 & 99.90 & 99.97 & 98.58 & 88.04 \\
        Pill       & 96.34 & 99.28 & 95.17 & 43.60 & 93.12 & 98.72 & 96.47 & 56.13 \\
        Screw      & 91.10 & 96.43 & 98.73 & 37.79 & 90.61 & 96.24 & 98.64 & 34.39 \\
        Tile       & 99.96 & 99.99 & 95.54 & 77.44 & 99.60 & 99.84 & 98.13 & 86.74 \\
        Toothbrush & 93.33 & 97.30 & 98.76 & 52.94 & 97.50 & 99.06 & 98.88 & 57.07 \\
        Transistor & 99.75 & 99.64 & 98.63 & 82.69 & 99.75 & 99.63 & 98.34 & 77.06 \\
        Wood       & 97.98 & 99.33 & 95.30 & 67.71 & 99.04 & 99.68 & 96.73 & 73.09 \\
        Zipper     & 99.03 & 99.77 & 97.42 & 59.04 & 99.87 & 99.97 & 97.61 & 59.89 \\
        \hline
        Mean       & {97.20} & {99.07} & {97.81} & {63.52} & \textbf{97.62} & \textbf{99.23} & \textbf{98.35} & \textbf{66.54} \\
        \hline
        \end{tabular}
    }
\end{table}

Combining cross-modal prompting and the segmentor (without Focal loss function) demonstrates synergy, achieving a pixel AUC of 98.05\% and pixel AP of 64.02\%, with notable enhancements in hazelnut (80.00\% vs. 72.92\%) and tile (85.43\% vs. 77.44\%). The full PromptMAD attains a pixel AUC of 98.35\% and pixel AP of 66.54\%, with improvements in metal\_nut (88.04\%), pill (56.13\%), hazelnut (82.45\%) and tile (86.74\%). Overall trends indicate that object classes benefit from the segmentor and Focal loss function for boundary precision, while textures gain from cross-modal prompting alignment. Isolated drops (e.g., in a grid for partial configurations) highlight the value of complete integration.

\begin{table}[]
    \centering
    \caption{Ablation study results on MVTec-AD dataset \cite{mvtec} focusing on Pixel AP, comparing our proposed method with baseline OneNIP \cite{onenip}, and individual components. Values in bold indicate improvement over the baseline OneNIP.}
    \vspace{1mm} 
    \label{tab:ablation}
    \resizebox{\linewidth}{!}{
        \begin{tabular}{|l|c|c|c|c|c|c|c|}
        \hline

        \multirow{2}{*}{Class} & \multirow{2}{*}{Baseline} & Only     & Only & VLM + segmentor & Only  & \multirow{2}{*}{PromptMAD} \\
                               &    & segmentor  & VLM  & (no Focal)    & Focal &  \\
        
        \hline
        Bottle & 82.62 & \textbf{84.22} & 80.44 & \textbf{83.18} & \textbf{83.09} & \textbf{85.13} \\
        Cable & 63.48 & \textbf{64.16} & \textbf{64.74} & 62.94 & 61.22 & \textbf{65.01} \\
        Capsule & 50.06 & 44.73 & \textbf{51.47} & 49.23 & 45.86 & 48.88 \\
        Carpet & 68.39 & 67.93 & \textbf{68.58} & \textbf{70.05} & 68.21 & \textbf{70.31} \\
        Grid & 45.43 & 42.07 & 44.46 & 38.46 & 43.94 & 42.31 \\
        Hazelnut & 72.92 & \textbf{76.31} & 69.92 & \textbf{80.00} & \textbf{75.53} & \textbf{82.45} \\
        Leather & 71.07 & \textbf{72.08} & \textbf{72.45} & 71.03 & 70.58 & \textbf{71.62} \\
        Metal Nut & 77.66 & \textbf{77.81} & 73.24 & 76.98 & \textbf{79.29} & \textbf{88.04} \\
        Pill & 43.60 & 34.86 & 42.21 & 43.25 & 35.68 & \textbf{56.13} \\
        Screw & 37.79 & 26.97 & 37.21 & \textbf{39.45} & 29.45 & 34.39 \\
        Tile & 77.44 & 73.37 & 76.47 & \textbf{85.43} & 76.45 & \textbf{86.74} \\
        Toothbrush & 52.94 & {59.76} & 51.93 & \textbf{54.36} & \textbf{54.10} & \textbf{57.07} \\
        Transistor & 82.69 & 80.11 & 82.11 & 79.93 & 81.43 & 77.06 \\
        Wood & 67.71 & \textbf{69.19} & 62.76 & \textbf{70.65} & \textbf{71.23} & \textbf{73.09} \\
        Zipper & 59.04 & 58.59 & \textbf{59.36} & 55.30 & 56.74 & \textbf{59.89} \\
        \hline
        Mean & 63.52 & 62.14 & 62.49 & \textbf{64.02} & 62.19 & \textbf{66.54} \\
        \hline
        \end{tabular}
    }
\end{table}

\section{CONCLUSION}
\label{sec:conclusion}

In this work, we introduce PromptMAD, a cross-modal prompting framework that significantly advances unsupervised multi-class visual anomaly detection by integrating semantic text prompts via CLIP for enriched reconstruction, Focal Loss for imbalance mitigation, and a supervised segmentor fusing multi-scale convolutions, Transformer attention, and diffusion refinement for high-precision localization. Results yield substantial improvements: mean pixel AUC from 97.81\% to 98.35\% and pixel AP from 63.52\% to 66.54\%, with pronounced gains in textured classes like metal\_nut and pill. This establishes new benchmarks in unified AD while maintaining efficiency, hypothesizing broader applicability to diverse domains.

\vfill\pagebreak


\begingroup
\setlength{\itemsep}{0pt}
\setlength{\parskip}{0pt}
\setlength{\parsep}{0pt}
\bibliographystyle{IEEEbib}
\bibliography{refs}
\endgroup

\end{document}